\documentclass[runningheads]{llncs}
\usepackage[T1]{fontenc}
\usepackage{graphicx}
\usepackage{times}  
\usepackage{helvet}  
\usepackage{courier}  
\usepackage[hyphens]{url}  
\usepackage{graphicx} 
\usepackage{caption} 
\usepackage{amsmath} 
\usepackage{amssymb}
\usepackage{bm}
\usepackage{array,multirow,graphicx}
 \usepackage{float}
\DeclareCaptionStyle{ruled}{labelfont=normalfont,labelsep=colon,strut=off} 
\usepackage{algorithm}
\usepackage{algorithmic}

\usepackage{newfloat}
\usepackage{listings}
\floatstyle{ruled}
\newfloat{listing}{tb}{lst}{}
\floatname{listing}{Listing}

\usepackage{array}
\newcolumntype{x}[1]{>{\centering\arraybackslash\hspace{0pt}}p{#1}}

\begin{document}
\title{DeepPAMM: Deep Piecewise Exponential Additive Mixed Models for Complex Hazard Structures in Survival Analysis}
\titlerunning{Deep Piecewise Exponential Additive Mixed Models}
%

\author{
    Philipp Kopper\textsuperscript{1} \and Simon Wiegrebe \and Bernd Bischl \and Andreas Bender\textsuperscript{2}\and David R{\"u}gamer\textsuperscript{2}
}

\institute{Department of Statistics, LMU Munich, Ludwigstr. 33, 80539 Munich (Germany) \\
 \email{\{firstname\}.\{lastname\}@stat.uni-muenchen.de}}

\authorrunning{P. Kopper et al.}
%
%
\maketitle              
\begin{abstract}
Survival analysis (SA) is an active field of research that is concerned with time-to-event outcomes and is prevalent in many domains, particularly biomedical applications. Despite its importance, SA remains challenging due to small-scale data sets and complex outcome distributions, concealed by truncation and censoring processes. The piecewise exponential additive mixed model (PAMM) is a model class addressing many of these challenges, yet PAMMs are not applicable in high-dimensional feature settings or in the case of unstructured or multimodal data. We unify existing approaches by proposing DeepPAMM, a versatile deep learning framework that is well-founded from a statistical point of view, yet with enough flexibility for modeling complex hazard structures. We illustrate that DeepPAMM is competitive with other machine learning approaches with respect to predictive performance while maintaining interpretability through benchmark experiments and an extended case study.
\end{abstract}
\section{Introduction}
\footnotetext[1]{Corresponding author.}
\footnotetext[2]{Contributed equally.}
Deep learning (DL) excels in many different areas of application through flexible and versatile network architectures. This has also been demonstrated in survival analysis (SA) \cite{ranganath.deep.2016,lee.deephit.2018}, where it is often not straightforward to apply off-the-shelf machine learning models. Apart from medical applications such as the prediction of time-to-death 
or the time to disease onset, 
time-to-event models are also applied in a variety of other domains. Among other fields, SA is successfully employed for predictive maintenance 
credit scoring 
, and customer churn prediction. 
%
In practice, time-to-event outcomes are not necessarily observed fully but might be censored, truncated or stem from a competing risks, or a multi-state process. 
While these aspects relate to the nature of the observation of event times, SA is also challenging due to the typically small amount of observations as well as complex feature effects and dependencies between observations. 
Medical survival data for instance potentially includes patient data of certain cohorts (such as patients from different hospitals with varying levels of patient care), longitudinal data with recurrent events or includes time-varying features such as a patient's vital status. Additionally, data can be multimodal (e.g., tabular patient information paired with medical images). 



\paragraph{Our contribution}

In this paper, we introduce a novel method called \emph{DeepPAMM} for continuous time-to-event data that enables the hazard-based learning of survival models via neural networks and supports
    1) many common survival tasks, including right-censored, left-truncated, competing risks, or multi-state data as well as recurrent events; 
    2) the estimation of inherently interpretable feature effects;
    3) learning from multiple data sources (e.g., tabular and imaging data); 
    4) time-varying effects and time-varying features;
    5) the modeling of repeated or correlated data using random effects.


\section{Related Literature} \label{sec:related}

Various models have been brought forward in SA. We will distinguish between models developed from a statistical point of view (Section~\ref{sec:refstats}), machine learning approaches (Section~\ref{sec:refml}) and recently proposed deep learning frameworks (Section~\ref{sec:refdl}).

\subsection{Piecewise exponential additive models and Cox proportional hazard models} \label{sec:refstats}

The Cox proportional hazard model (CPH) \cite{cox.regression.1972} is the most widely used survival model.
Under certain assumptions \cite{Whitehead1980} the Cox PH model is equivalent to the piecewise exponential model (PEM). The original formulation of the PEM, a parametric, linear effects, PH model, goes back to \cite{Friedman1982}. The general idea is to partition the follow-up time into $J$ intervals 
and to assume piecewise constant hazards in each interval. The originally proposed PEM requires a careful choice of the number and placement of interval cut-points. The piecewise exponential additive model (PAM) \cite{argyropoulos.analysis.2015,Bender2018b,Cai2002} is an extension of the PEM.  
PAMs estimate the baseline hazard and other time-dependent effects as smooth functions over time via penalized splines. This leads to more plausible and robust hazard estimates and (indirectly) lower computational cost. 
PAMs can be further generalized to piecewise exponential additive mixed models (PAMMs) by adding frailty terms (random effects). While PEMs and PAMMs can deal with many types of survival data (see, e.g., \cite{bender2018a,bender.general.2021}), they are limited w.r.t. the complexity of feature effects that they can estimate, especially in the case of high-dimensional features and interactions and cannot handle unstructured data.

\subsection{Machine Learning Approaches} \label{sec:refml}

In recent years a large number of machine learning methods for SA have been put forward. Random forest (RF) based methods include the random survival forest (RSF) \cite{ishwaran.random.2008} and more recently the oblique random survival forests (ORSF) \cite{jaeger.oblique.2019}.
In contrast to conventional RFs \cite{breiman2001random}, these adaptions make the models applicable to survival data by adjusting the splitting criterion. Next to trees and forests, several boosting methods exist, such as XGBoost \cite{chen.xgboost:.2016} or component-wise boosting for accelerated-failure time models \cite{schmid2008flexible}
and non-parametric hazard boosting \cite{lee.theory.2021}.
More recently and closest to our work, \cite{bender.general.2021} have proposed a general machine learning approach for various survival tasks based on PEMs and demonstrated its application using the standard XGBoost implementation.

\subsection{Deep Learning Approaches} \label{sec:refdl}

Various deep learning approaches have been proposed for SA, with the first approaches dating back to the mid-1990s (see, e.g., \cite{faraggi.neural.1995}). More recent approaches include both discrete-time methods like DeepHit \cite{lee.deephit.2018} 
or Nnet-survival \cite{gensheimer2019scalable} 
and continuous-time methods such as DeepSurv \cite{katzman2018deepsurv} or CoxTime \cite{kvamme2019time}.
DeepHit parametrizes the probability mass function by a neural network and specifically targets competing risks, but is only able to predict survival probabilities for a given set of discrete follow-up time points due to its time-discretization approach. Nnet-survival, by contrast, models discrete hazards and provides flexibility in terms of architecture choice, but it also relies on discretization of event times. DeepSurv is a Cox PH model with the linear predictor replaced by a deep feed-forward neural network. CoxTime further improves upon DeepSurv by allowing for time-varying effects, thereby overcoming the proportional hazards assumption. 
A deep Gaussian process to predict competing risks is proposed in \cite{alaa.2017}. While all previous methods focus on tabular data, a few multimodal networks such as \cite{haarburger2019image,kopper2021semi,polsterl2019wide,vale2021long} have also been proposed as well as survival tasks combined with a generative appraoch \cite{weber2021modelling}. 
The first combination of PEMs with a NN was proposed by \cite{liestbl.survival.1994}. 
\cite{biganzoli.general.2002} discussed the estimation of PEM by representing generalized linear models via feed-forward NNs, and \cite{fornili.piecewise.2014} proposed the estimation of the shape of the hazard rate with NNs. 
\cite{kvamme.continuous.2019} also discussed the parametrization of the PEM via NNs with application to tabular data. 
As for PEMs, the choice of cut-points in their framework is crucial for performance and computational complexity. Our framework eliminates this problem. 

\section{Piecewise Exponential Additive Models} \label{sec:PAM}

Survival analysis aims to estimate the survival function $S(t) = P(T>t)$. 
Instead of directly estimating $S(t)$, the hazard function
\begin{equation}\label{eq:hazard}
h(t) := \lim \limits_{\Delta t \to 0^+} \frac{P(t<T<t+\Delta t|T\geq t)}{\Delta t}
\end{equation}
is modeled. The survival function can be derived from $h(t)$ via $S(t) = \exp(-\int_{0}^{t}h(s)\,ds)$.
A hazard for time point $t\in \mathcal{T}$, conditional on a potentially time-varying feature vector $\bm{x}(t) \in \mathbb{R}^{P}$, can be defined by
\begin{equation}\label{eq:general.hazard}
h(t|\bm{x}(t),k) = \exp\left(\rho(\bm{x}(t),t,k)\right), k=1,\ldots,K.
\end{equation}
The function $\rho(\cdot)$ represents the effect of (time-dependent) features $\bm{x}(t)$ on the hazard and can itself be potentially time- and transition-specific. $k$ indicates a transition, e.g., from status $0$ to status $k$ in competing risks or the transition between two states in the multi-state setting.
In the following, we will set $K$ to 1 for better readability and only address the single risk application if not stated otherwise.
Further omitting the dependence on $t$, \eqref{eq:general.hazard} reduces to the familiar PH form known from the Cox model.

\subsection{Data Transformation}\label{ssec:ped}
PEMs and PAMs approximate \eqref{eq:general.hazard} via piecewise constant hazards, which requires a specific data transformation, creating one row in the data set for each interval a subject was at risk.
Assume observations (subjects) $i=1,\ldots,n$, for which the tuple $(t_i,\delta_i,\bm{x}_i)$ with event time $t_i$, event indicator $\delta_i \in \{0,1\}$ (1=event, 0=censoring) and feature vector $\bm{x}_i$ is observed. 
PAMs partition the follow up into $J$ intervals $(\kappa_{j-1},\kappa_j],\ j=1,\ldots,J$.  
This implies a new status variable $\delta_{ij} = 1$ if $t_{i} \in (\kappa_{j-1}, \kappa_j] \wedge \delta_{i} = 1$, and $0$ otherwise, indicating the status of subject $i$ in interval $j$. Further, we create a variable $t_{ij}$, the time subject $i$ was at risk in interval $j$, which will enter the analysis as an offset. 
Lastly, the variable $t_j$, (e.g., $t_j:=\kappa_j$) is a representation of time in interval $j$ and the feature based on which the model estimates the baseline hazard and time-varying effects. 
In order to transform the data to the piecewise exponential data format (PED), time-constant features $\bm{x}_{i}$ are repeated for each of $J_i$ rows, where $J_i$, denotes the number of intervals in which subject $i$ was at risk. This data augmentation step transforms a survival task into a standard Poisson regression task. 
Depending on the setting, e.g., right-censoring, recurrent events, left truncation, etc., the specifics of the data transformation vary, but the general principles remain the same.
For more details we refer to \cite{pammtools,bender.general.2021,ramjith_recurrent_2021}.

\subsection{Model Estimation}\label{ssec:pam_estimation}
Given the transformed data, PAMs approximate \eqref{eq:general.hazard} by
    $h(t|\bm{x}_i(t)) = \exp(\rho(\bm{x}_{ij},t_j)):=h_{ij}, \forall t \in (\kappa_{j-1},\kappa_j]$ ,
where $\bm{x}_{ij}$ is the feature vector of subject $i$ in interval $j$. Assuming $\delta_{ij}\sim \mathrm{Poisson}(\mu_{ij}=h_{ij}t_{ij})$, the log-likelihood contribution of subject $i$ is given by 
$\ell_i = \sum_{j=1}^{J_i}(\delta_{ij}\log(h_{ij}) - h_{ij}t_{ij})$,
where
\begin{equation*}\label{eq:pam.eta}
\log(h_{ij}) = \beta_0 + f_0(t_j) + \sum_{p=1}^P x_{ij,p} \beta_p + \sum_{l=1}^L f_l(x_{ij,l}),
\end{equation*}
with log-baseline hazard $\beta_0 + f_0(t_j)$, linear feature effects $\beta_p$ of features $x_{ij,p} \subseteq \bm{x}_{ij}$ and univariate, non-linear feature effects $f_l(x_{ij,l})$ of features $x_{ij,l}\subseteq\bm{x}_{ij}$.
Both $f_0$ and $f_l$ are defined via a basis representation, i.e., 
$f_l(x_{ij,l}) = \sum_{m=1}^{M_l}\theta_{l,m}B_{l,m}(x_{ij,l})$ with basis functions $B_{\cdot,m}(\cdot)$ (such as B-spline bases) and basis coefficients $\theta_{\cdot,m}$. To avoid underfitting, the basis dimensions $M_0$ (for $f_0$) and $M_l$ (for $f_l$) are set relatively high. To avoid overfitting, the basis coefficients are estimated by optimizing an objective function that penalizes differences between neighboring coefficients. Let $\boldsymbol{\beta} = (\beta_0,\ldots,\beta_P)^\top$ and $\boldsymbol{\theta}_{l}=(\theta_{l,1},\ldots, \theta_{l,M_l})^\top$, $l=0,\ldots,L$. The objective function minimized to estimate PAMs is the penalized negative log-likelihood given by
$- \log \mathcal{L}(\boldsymbol{\beta},\boldsymbol{\theta}_0,\ldots,\boldsymbol{\theta}_L) + \sum_{l=0}^{L} \psi_{l}\Psi(\boldsymbol{\theta}_{l})$,
where the first term is the standard negative logarithmic Poisson likelihood, comprised of likelihood contributions $\ell_i$, and the second term $\Psi(\boldsymbol{\theta}_{l})$ is a quadratic penalty with smoothing parameter $\psi_{l}\geq 0$ for the respective spline $f_l$. Larger $\psi_{l}$ lead to smoother $f_l$ estimates (see \cite{wood.generalized.2017,bender2018a} for details). 

\section{Deep Piecewise Exponential Additive Mixed Models} \label{sec:deeppam}
\begin{figure*}[t]
\small
    \centering
    \includegraphics[width = 0.99\textwidth]{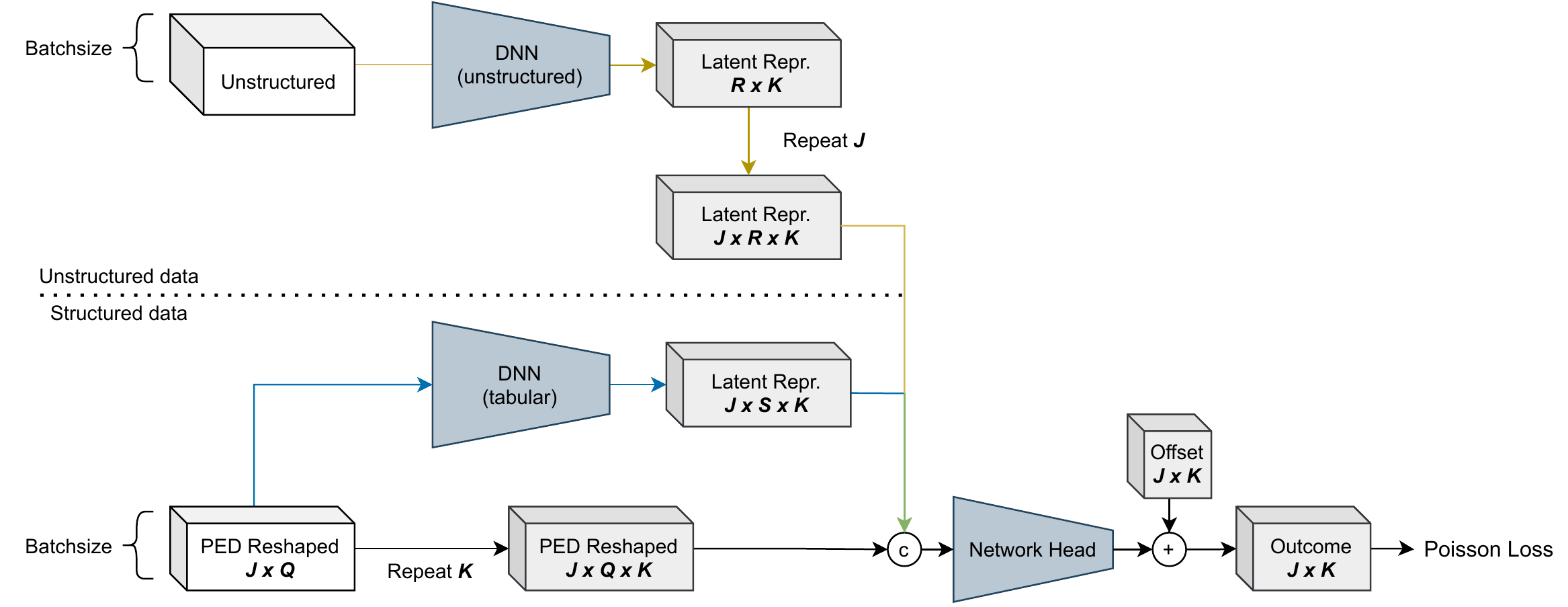}
    \caption{Exemplary architecture of a DeepPAMM. A DeepPAMM comprises a PAMM (black path) and additionally either a deep neural network (DNN) for unstructured data (yellow path), a DNN for tabular features (blue path), or both. The unstructured data, e.g., images, are summarized to latent representations of size $R$, repeated $J$ times, and concatenated (c) with the tabular data's latent representation of size $S$, as well as raw tabular data of size $Q$. Finally, the offset is added to the output and the network is trained using the Poisson loss for each of the $K$ competing risks. }
    \label{fig:deepPAM}
\end{figure*}
DeepPAMMs extend PAM(M)s with hazard as defined in \eqref{eq:general.hazard} by allowing for deep neural networks (NN) in the additive predictor. Instead of combining PAMMs with (deep) NNs in a two-stage approach, we embed the PAMM into the NN similar to \cite{Ruegamer.2020} and train the network based on the (penalized) likelihood in an end-to-end manner. 

\paragraph{Network Definition}

While PAMMs restrict $\rho$ to structured additive effects, the hypothesis space of DeepPAMMs can also be modeled using a deep NN. 
Assume that the NN $d(\cdot)$ is used to process a potentially time-varying (unstructured) data source $\bm{z}(t)$. 
We first assume a time-constant effect of $\bm{z}(t)$ and extend the PAMM's definition to
\begin{equation}
    \label{eq:deeppem.1}
    h(t|\bm{x}(t),\bm{z}(t)) = \exp\bigl\{ \rho(\bm{x}(t),t) + d(\bm{z}(t))\bigr\},
\end{equation} 
by adding one (or several) NN predictor(s) to the structured predictor. 

The predictor $d(\bm{z}(t))$ can be modeled using an arbitrary NN.
For example, a DeepPAMM can combine a PAM with an additional NN to explore non-linearities and interactions in tabular features (beyond the ones specified in the structured part).
Alternatively, a DeepPAMM can combine different data modalities, e.g., tabular patient data and corresponding medical scans using a convolutional NN for $d$.
By \eqref{eq:deeppem.1}, DeepPAMM learns a piecewise constant hazard rate
\begin{equation} \label{eq:deeppem.2}
h_{ij} = \exp\bigl\{\bm{B}_{ij}\bm{w} + \sum_{u=1}^U \zeta_{ij,u} \gamma_u \bigr\},
\end{equation}
for each observation $i$ and each discrete interval $j$, where $\bm{B}_{ij}$ subsumes all $Q$ structured features (linear and basis evaluated features) with weights $\bm{w}$.
$\zeta_{ij,1},\ldots,\zeta_{ij,U}$ are $U = R + S$ latent representations learned from the deep network part that processes tabular data (into $S$ latent features) and unstructured data (into $R$ latent features). The network then combines these $U$ latent representations to learn the effect $\gamma_1,\ldots,\gamma_U$ of each of these feature effects. Due to the additive structure in predictor \eqref{eq:deeppem.2}, the structured terms with linear effects $\bm{w}$ preserve their interpretability inherited from PAMMs. 

\paragraph{PED and Latent Representations}

$d(\bm{z}(t))$ can be viewed as linear effects of $U$ latent representations derived from inputs $\bm{z}(t)$.
In \eqref{eq:deeppem.1} this representation is combined with the structured features in a last layer summing up the two predictors.
If $\bm{z}$ is constant over time, i.e., $\bm{z}(t) \equiv \bm{z}$, it is not straightforward to combine these latent representations with the PED format properly.
A naive approach would be to repeat the {original} data source $\bm{z}$ over all $J$ intervals. 
This, however, leads to significant computational overhead and storage of redundant information. Instead, we resort to weight-sharing and reshaping within the network that allows learning a single latent representation per observation for all $J$ intervals (cf. Figure~\ref{fig:deepPAM}). First, the original tabular data is transformed to the PED format prior to the network training. Subsequently, the reshaped three-dimensional PED tensor batches with the same sampling dimension as the unstructured data source $\bm{z}$ are passed through the network. $\bm{z}$ itself is transformed into $R$ latent representations and then repeated $J$ times for each interval. This avoids repeating the original unstructured data source multiple times. Finally, we combine these representations with the original tabular data and the $S$ non-linear representations of the structured data part into a joint set of features.  
While we here focus on time-constant unstructured data, our framework can be extended to allow for time-varying unstructured features by simply also supplying the time $t$ to the deep NN $d$ explicitly, i.e.,
extending $d(\bm{z}(t))$ in \eqref{eq:deeppem.1} to $d(\bm{z}(t), t)$.

\paragraph{Learning Non-proportional Hazards} \label{sec:nonprop}

PAMMs allow for non-proportional hazards via an interaction of features $\bm{x}$ with a feature that represents time in each of the J intervals.
In practice, however, the accompanying computational complexity and manual definition of these interactions are often infeasible. 
In DeepPAMM, such interactions can be modeled using an appropriate multilayer NN architecture. In particular, interactions between features $\bm{z}(t)$ and the follow-up $t$ can be expressed by
    $h(t|\bm{x}(t),\bm{z}(t)) = \exp\bigl\{\rho\left(\bm{x}(t),d(\bm{z}(t)),t\right)\bigr\}$,
where $\rho$ now also depends on the the specified NN to model a non-proportional hazard in $\bm{z}(t)$.
As the PH assumption is a helpful inductive bias for applications with small sample sizes, we recommend this extension for larger data sets or in applications where the PH assumption is clearly violated.

\paragraph{Learning Competing Risks Hazards}

When modeling competing risks data with $K$ different risks that determine the time-to-event, one is interested in retrieving the cumulative incidence functions of each risk (CIFs).
Our architecture allows for a holistic way of modeling the hazard of subject $i$ in interval $j$ and cause $k$ in a joint NN:
    $h_{ijk} = \exp\bigl\{\bm{B}_{ijk}\bm{w}_k + \sum_{u=1}^U \zeta_{ijk,u} \gamma_{k,u} \bigr\}$,
where $\bm{B}_{ijk}$ is equivalent to the input $\bm{B}_{ij}$, i.e., we repeat $\bm{B}_{ij}$ $K$ times so that cause-specific weights $\bm{w}_k$ share the same inputs. Similarly, the latent representations $\zeta_{ijk,u}$ now also depend on the risk $k=1,\ldots,K$ to yield cause-specific effects $\gamma_{k,u}$ for each latent feature. Figure~\ref{fig:deepPAM} illustrates the CR case for an exemplary network architecture. Training the network is based on a joint loss summing up all $K$ loss contributions for each CR and weighted by binary interval weights if the observation is still at risk in the $j$th interval and $0$ if not.

\paragraph{Learning Mixed Effects and Recurrent Events}

In many SA settings, data comes in clusters.
For example, the survival of patients has been observed at different locations.
This is typically the case for multi-center studies for which survival may substantially vary between clusters while being more homogeneous within each cluster.
A random effect (RE), i.e., a linear effect for each cluster with a normal prior, can account for this within-cluster correlation.
REs can also be used to account for repeated measurements and recurrent events.
Optimization of NNs with random or mixed effects can be done using an EM-type optimization routine (see, e.g., \cite{Xiong.2019}), by training a Bayesian NN (see, e.g., \cite{Hernandez.2015}), or by tuning the prior variance based on the equivalence of a random normal prior and a ridge-penalized effect (see, e.g., \cite{wood.generalized.2017}).
While learning the RE prior variance explicitly is desirable, a carefully chosen ridge penalization should yield similar results (due to their mathematical equivalence) while being more straightforward to incorporate in most NNs. 

\section{Numerical Experiments} \label{sec:exp}

We first explore DeepPAMM by investigating some of the proposed model properties in a simulation study.
Additionally, we compare DeepPAMM with state-of-the-art algorithms on various benchmark data sets including real-world medical applications.
We examine model performance via the integrated Brier score (IBS) \cite{graf1999assessment}, which measures both, discrimination and calibration of predicted survival probabilities. 
Instead of integrating over the whole time domain, we evaluate the IBS at the first three quartiles (Q25, Q50, Q75) of the observed event times in the test set, in order to assess the performance at different time points.
While DL-based approaches usually require large data sets for training, DeepPAMM also works well in small data set regimes.
In the worst case, if there is not enough data to train the deep part of our network, the structured network part will dominate the predictions. 
DeepPAMM will then effectively fall back to estimate a PAMM, which in turn is well suited for small data sets.
This property is especially important in SA where most data sets are relatively small.

\subsection{Simulation and Ablation Study}

\renewcommand{\arraystretch}{1.2}

\begin{table}[t]
\small
 \caption{Comparison of the average IBS (with standard deviation in brackets) across the three quartiles Q25, Q50, Q75 (rows) for different methods (columns) in different study settings. The $\dagger$-symbol indicates methods that can only take tabular data information into account.} 
  \label{tab:simstudy} 
\begin{tabular}{m{30pt}|x{23.5pt}x{23.5pt}x{23.5pt}|x{23.5pt}x{23.5pt}x{23.5pt}|x{23.5pt}x{23.5pt}x{23.5pt}|x{23.5pt}x{23.5pt}x{23.5pt}}
        & \multicolumn{3}{c|}{CR (cause 1)}       & \multicolumn{3}{c|}{CR (cause 2)}       & \multicolumn{3}{c|}{Mixed effects}      & \multicolumn{3}{c}{Multimodal}          \\
        & Q25         & Q50         & Q75         & Q25         & Q50         & Q75         & Q25         & Q50         & Q75         & Q25         & Q50         & Q75         \\ \hline
KM      & 5.1 (0.43) & 10.8 (0.49) & 16.2 (0.57)  & 4.3 (0.31)) & 9.1 (0.52)   & 13.6 (0.60)   & 7.0 {(0.47)}  & 13.5 {(0.57)}  & 18.2 {(0.68)} & 4.1${}^\dagger$ (0.44) & 8.0${}^\dagger$ (0.49)  & 12.1${}^\dagger$ (0.66)   \\
PAMM & 3.3 (0.32) &    6.0 (0.52)         &   8.9 (0.52)          &    2.5 (0.21)          &     4.5 (0.38)   &    7.0 (0.61)   &   3.9 (0.49)  &    6.9 (0.71)    &   9.1 (0.89)       &      3.7${}^\dagger$ (0.43) &   6.3${}^\dagger$ (0.63)  & 8.6${}^\dagger$ (0.71)   \\
Ours    &  \textbf{2.9} (0.41)           &    \textbf{5.4} (0.40)         &      \textbf{8.1} (0.43)       &      \textbf{2.4} (0.38)       &     \textbf{4.4} (0.46)         &   \textbf{6.8} (0.68)       &       \textbf{3.2} (0.41)       &   \textbf{5.7} (0.61)           &     \textbf{7.5} (0.69)        &     \textbf{3.6} (0.43)        &    \textbf{6.1} (0.52)          &   \textbf{8.4} (0.65)          \\ \hline
Optimal &    {2.9} (0.72)         &     {5.4} (0.80)        &      {8.0} (0.79)       &       {2.1} (0.22)      &      {4.1} (0.39)        &  {6.5} (0.63)           &   {2.9} (0.36)          &     {5.2} (0.57)        &    {6.8} (0.68)         &      {3.6} (0.42)       &    {6.1} (0.58)         & {8.3} (0.68)            
\end{tabular}
\end{table}

\renewcommand{\arraystretch}{1}

The goal of our simulation study is to investigate the performance of DeepPAMM under various controlled settings with a focus on 1) mixed effects, 2) competing risks, 3) multimodal data. For all simulations, the data generating process incorporates both, linear effects and non-linear interactions. For every setting, we repeat this procedure 25 times to account for variance in data generation and model fitting. In the spirit of an ablation study, we compare DeepPAMM with its corresponding PAM(M) to investigate the attribution of performance gains as well as the relation to an ideal model (Optimal). 

For \textbf{competing risks}, we simulate two competing risks based on two different hazards structures. 
While cause 1 is based on 5 features and multiple non-linear interaction effects, cause 2 relates to 3 features and a more moderate level of interactions as well as non-linearities. 

For \textbf{mixed effects}, we simulate repeated measurements by defining 60 clusters and drawing a random effect for each cluster unit from a normal distribution with zero mean and a standard deviation of 1.5. 
Before training DeepPAMM, we pre-train the random effects of the DeepPAMM with the corresponding PAMM and use the associated ridge penalty as a warm start for tuning.

For the \textbf{multimodal data} scenario, we simulate log-hazards based on linear latent effects from point clouds (PC) based on the data set from ModelNet10 \cite{wu20153d}.
Each of the PC labels is associated with a different latent coefficient ranging from -0.5 to 0.75. The hazard is defined to depend on these latent coefficients as well as on tabular features. A reduced PointNet \cite{qi2017pointnet} is used to model the PCs.
This set up has been adapted from \cite{kopper2021semi}.


\paragraph{Results}
Model comparisons are provided in Table~\ref{tab:simstudy}. In summary, our proposed model is the best performing method across all three settings and in most cases yields performance values close to the optimal error in terms of the IBS.
While performance gains in absolute terms seem small, the decrease in IBS relative to the optimal error is especially noteworthy for CR (cause 1) and the mixed effects setting.
Results confirm that DeepPAMM works well in various of the proposed data situations. The ablation study further justifies the deep part of DeepPAMM by its improved performance in comparison to PAMM. 

\subsection{Benchmark Analysis}

We compare our approach with various state-of-the-art methods (Table~\ref{tab:resultsBM}).
Comparisons include a tree-based method (ORSF; \cite{jaeger.oblique.2019}), a boosting approach (PEMXGB; \cite{bender.general.2021}), as well as (DeepHit; \cite{lee.deephit.2018}), a well-established deep NN for SA.
As baseline models we use a Kaplan-Meier estimator (KM; \cite{cox.regression.1972}) and a Cox PH model (CPH; \cite{cox.regression.1972}). We restrict our comparison to directly and publicly available SA data sets that have been used in the benchmarks of methods listed above, namely
 \emph{tumor} \cite{pammtools}, \emph{gbsg2} \cite{schumacher1994randomized},
 \emph{metabric} (cf. \cite{lee.deephit.2018}),
 \emph{breast} \cite{ternes.identification.2017},
 \emph{mgus2} \cite{kyle2002long}, and 
 \emph{icu} (cf. \cite{hartl2019calorie}).
For each method, we perform a random search with 50 configurations and compare the aggregated (mean and std. deviation) test set performances on 25 distinct train-test-splits.
The data sets impose different challenges, including CR (icu, mgus2), high-dimensional data (breast), and mixed effects (icu).
For these, DeepPAMM is consistently among the best-performing survival models. The main point here is that DeepPAMM is competitive compared to other state-of-the-art methods while maintaining interpretability as illustrated in \ref{sec:case-study}.
\begin{table}[t]
\small
    \centering
          \caption{Performance comparison based on the IBS ($\downarrow$) at the three quartiles (Q25, Q50, Q75) across different data sets (rows) and models (columns) with best models per row highlighted in bold. Missing entries are due to missing support for CRs.} 
  \label{tab:resultsBM} 
\begin{tabular}{@{\extracolsep{1pt}} cccccccc} 
\hline \\[-1.8ex] 
Data Set &  & KM & Cox PH & ORSF & PEMXGB & DeepHit & DeepPAMM \\ 
\hline \\[-1.8ex] 
& Q25 & 6.6 (0.59) & 6.0 (0.58) & \textbf{5.5} (0.56) & 5.7 (0.63) & 5.6 (0.55) & 5.7 (0.59)  \\
tumor & Q50 & 12.3 (0.86)  & 11.2 (0.82) & \textbf{10.8} (0.91) & 10.9 (1.05) & 11.0 (0.96) & 10.9 (0.86) \\
& Q75 & 17.6 (0.79) & 16.3 (0.77) & 16.3 (0.85) & \textbf{16.2} (0.92)  & 16.4 (0.95) & \textbf{16.2} (0.81)  \\
\hline \\[-1.8ex] 
& Q25 & 3.1 (0.49) & 3.1 (0.45) & \textbf{3.0} (0.45) &  \textbf{3.0} (0.46) & 3.1 (0.49) & 3.1 (0.41) \\
gbsg2 & Q50 & 6.8 (0.80) & 6.5 (0.72) & \textbf{6.2} (0.70) & 6.3 (0.68) & 6.6 (0.8) & 6.5 (0.69)  \\
& Q75 & 12.5 (1.04) & 11.4 (0.94) & \textbf{11.1} (0.95) & 11.3 (1.01)  & 11.9 (0.99) & 11.5 (0.95) \\
\hline \\[-1.8ex] 
& Q25 & 4.0 (0.22) & 4.0 (0.26) & 4.1 (0.25) & \textbf{3.8} (0.22) & 4.0 (0.22) & 3.9 (0.27) \\
metabric & Q50 & 8.6 (0.51) & 8.2 (0.54)  & 8.9 (0.46) & \textbf{7.8} (0.49) & 8.4 (0.45) & 7.9 (0.45) \\
& Q75 & 14.0 (0.38) & 12.9 (0.47) & 14.7 (1.19) & \textbf{12.3} (0.51) & 13.5 (0.40)  & 12.6 (0.44)  \\
\hline \\[-1.8ex] 
& Q25 & \textbf{1.9} (0.61) & - & 2.0 (0.59) & 2.0 (0.57) & 2.1 (0.54) & \textbf{1.9} (0.60)\\
breast & Q50 & 4.1 (0.89) & - & \textbf{4.0} (0.80) & \textbf{4.0} (0.83) & 4.2 (0.81) & \textbf{4.0} (0.90)\\
& Q75 & 7.1 (1.13) & - & \textbf{6.7} (0.96) & \textbf{6.7} (1.10) & 7.1 (1.02) & 6.8 (1.33) \\
\hline \\[-1.8ex] 
& Q25 & \textbf{1.1} (0.21) & - & - &1.9 (0.34) & \textbf{1.1} (0.21)  & \textbf{1.1} (0.21) \\
mgus2 & Q50 & \textbf{2.2} (0.34) & - & - & 4.1 (0.55) & \textbf{2.2} (0.34) & \textbf{2.2} (0.34) \\
(cause 1) & Q75 & \textbf{3.4} (0.48) & - & - & 6.9 (0.69) & 3.5 (0.51) & \textbf{3.4} (0.49) \\
\hline \\[-1.8ex] 
& Q25 & 8.7 (0.52) & - & - & 8.6 (0.65)  & \textbf{8.1} (0.55) & 8.3 (0.49) \\
mgus2 & Q50 & 14.4 (0.61) & - & - & 13.9 (0.84)  & \textbf{12.9} (0.66) & 13.1 (0.65) \\
(cause 2) & Q75 & 18.4 (0.60) & - & - & 17.9 (1.04) & \textbf{15.8} (0.67) & 16.0 (0.67) \\
\hline \\[-1.8ex] 
& Q25 & \textbf{1.3} (0.06) & - & - & 1.4 (0.66) & \textbf{1.3} (0.06)  & \textbf{1.3} (0.06) \\
icu & Q50 & 3.6 (0.14) & - & - & 3.6 (0.13) & \textbf{3.5} (0.13) & \textbf{3.5} (0.13) \\
(cause 1) & Q75 & 6.7 (0.19) & - & - & 6.7 (0.19) & 6.5 (0.20) & \textbf{6.4} (0.19) \\
\hline \\[-1.8ex] 
& Q25 &  3.5 (0.15) & - & - & 3.5 (0.14)  & \textbf{3.4} (0.14) & \textbf{3.4} (0.14) \\
icu & Q50 & 7.6 (0.17) & - & - & 7.6 (0.17) & \textbf{7.3} (0.17) & \textbf{7.3} (0.16) \\
(cause 2) & Q75 & 12.0 (0.15)   & - & - & 12.1 (0.17) & 11.5 (0.20) & \textbf{11.3} (0.17) \\
\hline \\[-1.8ex] 
\end{tabular} 
\end{table}

\subsection{Extended Case Study}\label{sec:case-study}

\begin{table}[t]
\small
    \centering
          \caption{Performance comparison based on the IBS ($\downarrow$) at the three quartiles (Q25, Q50, Q75) across different models (columns) for the data set of \cite{taylor2017spatial} with best models per row highlighted in bold. The performance has been assessed using 25 train-test splits.} 
  \label{caseStudy} 
\begin{tabular}{@{\extracolsep{1pt}} cccccccc} 
\hline \\[-1.8ex] 
Quantile  & KM & PAMM & DeepPAMM  \\ 
\hline \\[-1.8ex] 
 Q25 & 12.8 (0.28) & 12.3 (0.30) & \textbf{12.2} (0.32)  \\
 Q50 & 18.1 (0.18)  & 16.9 (0.21) & \textbf{16.7} (0.23) \\
Q75 & 19.9 (0.11) & 18.4 (0.14) & \textbf{18.2} (0.14) \\
\hline \\[-1.8ex] 
\end{tabular} 
\end{table}

\begin{figure*}[t]
\small
    \centering
    \includegraphics[width = 0.95\textwidth]{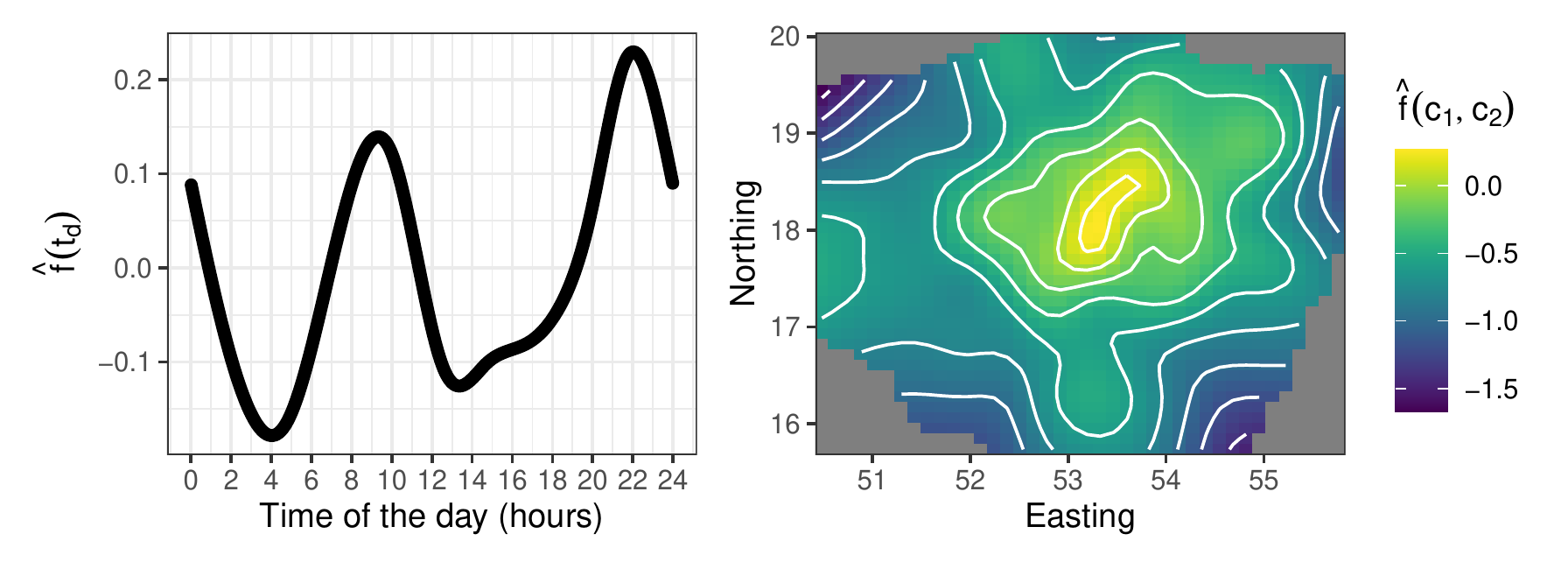}
    \caption{Smooth cyclic (left) and spatial (right) effect of a DeepPAMM. Effects are from a single of the 25 runs.}
    \label{fig:interpretable}
\end{figure*}

In this extended case study, we show how DeepPAMM can be used to obtain interpretable feature effects and at the same time incorporate potentially high-dimensional interactions. 
To illustrate this, we apply DeepPAMM to spatio-temporal data where the outcome is response times (time-to-arrival) of the London fire brigade to fire-related emergency calls \cite{taylor2017spatial}.
Additionally, the data includes geographic coordinates of the site of the fire as well as information about the ward from which the truck was deployed and the time of day of the incident.
We expect a non-linear effect of the time of day that varies with day and night times as well as traffic hours and a bivariate spatial effect of the location with different hazards in different regions of the city. Therefore, we model the hazard for arrival at time $t$ given time of day $t_d$, spatial coordinates ($c_1$ and $c_2$) and ward $v=1,\ldots,V$ as
\begin{align}
    \log(h(t|t_d, c_1, c_2, v)) = & \underbrace{\beta_0 + f_0(t) + f_1(t_d) + f_2(c_1, c_2) + b_{v}}_{\text{structured}}
    + \underbrace{d(t, t_d, c_1, c_2, v)}_{\text{unstructured}}\nonumber
\end{align}
where $f_1(t_d)$ is estimated as a cyclic spline that enforces equal values of the function at 0 and 24 hours, $f_2(c_1, c_2)$ is a bivariate tensor product spline and $b_{v}$ are random effects for the individual wards. In the unstructured part, we additionally allow for high-dimensional interactions between the features from the structured part. This way, we can investigate whether the predictive performance can be improved beyond the structured part. Structured effects are given in Figure~\ref{fig:interpretable}. For interpretation, note that higher hazards imply shorter response times, thus response times are on average longer during night hours and between 12 and 18 p.m. as well as in the periphery of the city. 
The results w.r.t. the predictive performance are shown in Table~\ref{caseStudy}, where we compare our model with a KM baseline and the respective PAMM. 
In addition to the PAMM specification, our model includes a NN with three layers (64, 32, 8 neurons) to model feature interactions. The results indicate that on average the performance improves slightly when the unstructured part is added. Given the resulting standard deviations, we conclude that the structured part is sufficient. Further, DeepPAMM's structured effects are in line with results presented in \cite{taylor2017spatial}. This shows the strength of DeepPAMM: maintaining interpretability of covariate effects as illustrated in Figure~\ref{fig:interpretable}, while also allowing the investigation of additional effects in the unstructured part.

\section{Concluding Remarks} \label{sec:discussion}

We present DeepPAMM, a novel semi-structured deep learning approach to survival analysis.
Our experiments demonstrate that our model has high predictive capacity and is capable of modeling diverse complex data associations. DeepPAMM allows to include non-linear and feature interaction effects in the model, can be used to model non-proportional hazards, time-varying effects and competing risks, while also accounting for correlation in the data using mixed effects. The deep part of the model further makes estimation in high-dimensional settings possible and can be used to include unstructured data into the survival analysis.
The additive predictor in our approach allows for straightforward interpretability and to recover the PAM(M) when no additional deep predictors are necessary. 
Our method can be fit using existing software solutions (e.g., deepregression \cite{rugamer2021deepregression}).

\section{Acknowledgements}

This work has been partly funded by the German Federal Ministry of Education and Research (BMBF) under Grant No. 01IS18036A. The authors of this work take full responsibility for its content.

\bibliographystyle{splncs04}
\bibliography{bibliography}

\end{document}